%% file: main.tex
\documentclass[10pt,twocolumn,letterpaper]{article}

\usepackage{cvpr}
\usepackage{times}
\usepackage{epsfig}
\usepackage{graphicx}
\usepackage{amsmath}
\usepackage{amssymb}
\usepackage{arydshln}
\usepackage{booktabs}
\usepackage{graphicx}
\usepackage{multirow}
\usepackage{tabu}
\usepackage{siunitx}
\usepackage[dvipsnames,svgnames,x11names]{xcolor}

\usepackage[pagebackref=true,breaklinks=true,letterpaper=true,colorlinks,bookmarks=false]{hyperref}

\cvprfinalcopy 

\def\cvprPaperID{322} 

\DeclareMathOperator{\sign}{sign}

\newcommand{\mycaption}[2]{\caption{\textbf{#1.}\xspace#2}}

\ifcvprfinal\fi
\begin{document}

\title{Self-Supervised Viewpoint Learning From Image Collections \vspace{-5mm}}

\author{
Siva Karthik Mustikovela\textsuperscript{1,2}\thanks{Siva Karthik Mustikovela was an intern at NVIDIA during the project.} \hspace{2mm}
Varun Jampani\textsuperscript{1} \hspace{2mm} Shalini De Mello\textsuperscript{1} \\
Sifei Liu\textsuperscript{1} \hspace{2mm}
Umar Iqbal\textsuperscript{1} \hspace{2mm}
Carsten Rother\textsuperscript{2} \hspace{2mm}
Jan Kautz\textsuperscript{1} \\
\vspace{1mm}
\textsuperscript{1}NVIDIA \qquad \textsuperscript{2}Heidelberg University\\ 
{\tt\small \{siva.mustikovela, carsten.rother\}@iwr.uni-heidelberg.de; varunjampani@gmail.com;} \\ 
\tt\small \{shalinig, sifeil, uiqbal, jkautz\}@nvidia.com}

\maketitle

\vspace{-6mm}

\input{00_abstract}
\input{01_introduction}
\input{02_relatedwork}
\input{03_method}

\input{04_results}
\input{05_conclusions}

{\small
\bibliographystyle{ieee_fullname}
\bibliography{SSV}
}

\clearpage

\input{06_supp}

\end{document}

%% file: 00_abstract.tex
\begin{abstract}
\vspace{-2mm}
Training deep neural networks to estimate the viewpoint of objects requires large labeled training datasets. However, manually labeling viewpoints is notoriously hard, error-prone, and time-consuming. On the other hand, it is relatively easy to mine many unlabelled images of an object category from the internet, e.g., of cars or faces. We seek to answer the research question of whether such unlabeled collections of in-the-wild images can be successfully utilized to train viewpoint estimation networks for general object categories purely via self-supervision. Self-supervision here refers to the fact that the only true supervisory signal that the network has is the input image itself. 
We propose a novel learning framework which incorporates an analysis-by-synthesis paradigm to reconstruct images in a viewpoint aware manner with a generative network, along with symmetry and adversarial constraints to successfully supervise our viewpoint estimation network. We show that our approach performs competitively to fully-supervised approaches for several object categories like human faces, cars, buses, and trains. Our work opens up further research in self-supervised viewpoint learning and serves as a robust baseline for it. We open-source our code at \url{https://github.com/NVlabs/SSV}.
\end{abstract}

%% file: 01_introduction.tex
\vspace{-5mm}
\section{Introduction}
\label{sec:intro}
\vspace{-1mm}

3D understanding of objects from 2D images is a fundamental computer vision problem. Object viewpoint (azimuth, elevation and tilt angles) estimation provides a pivotal link between 2D imagery and the corresponding 3D geometric understanding. In this work, we tackle the problem of object viewpoint estimation from a single image. Given its central role in 3D geometric understanding, viewpoint estimation is useful in several vision tasks such as object manipulation~\cite{xiang2018rss:posecnn}, 3D reconstruction~\cite{kundu20183d}, image synthesis~\cite{chen2019mono} to name a few. Estimating viewpoint from a single image is highly challenging due to the inherent ambiguity of 3D understanding from a 2D image. Learning-based approaches, \textit{e.g.},~\cite{LiaoCVPR19, grabner20183d, zhou2018starmap, mahendran20173d, su2015render, tulsiani2015viewpoints, gu2017dynamic, yang2019fsa}, using neural networks that leverage a large amount of annotated training data, have demonstrated impressive viewpoint estimation accuracy. A key requirement for such approaches is the availability of large-scale human annotated datasets, which is very difficult to obtain. A standard way to annotate viewpoints is by manually finding and aligning a rough morphable 3D or CAD model to images~\cite{fanelli2013random, zhu2017face, xiang2014beyond}, which is a tedious and slow process. This makes it challenging to create large-scale datasets with viewpoint annotations. Most existing works~\cite{grabner20183d, massaBMVC2016, su2015render, LiaoCVPR19, zhu2017face, gu2017dynamic} either rely on human-annotated viewpoints or augment real-world data with synthetic data. Some works~\cite{grabner20183d} also leverage CAD models during viewpoint inference.

\begin{figure}
	\centering
	\includegraphics[width=\linewidth]{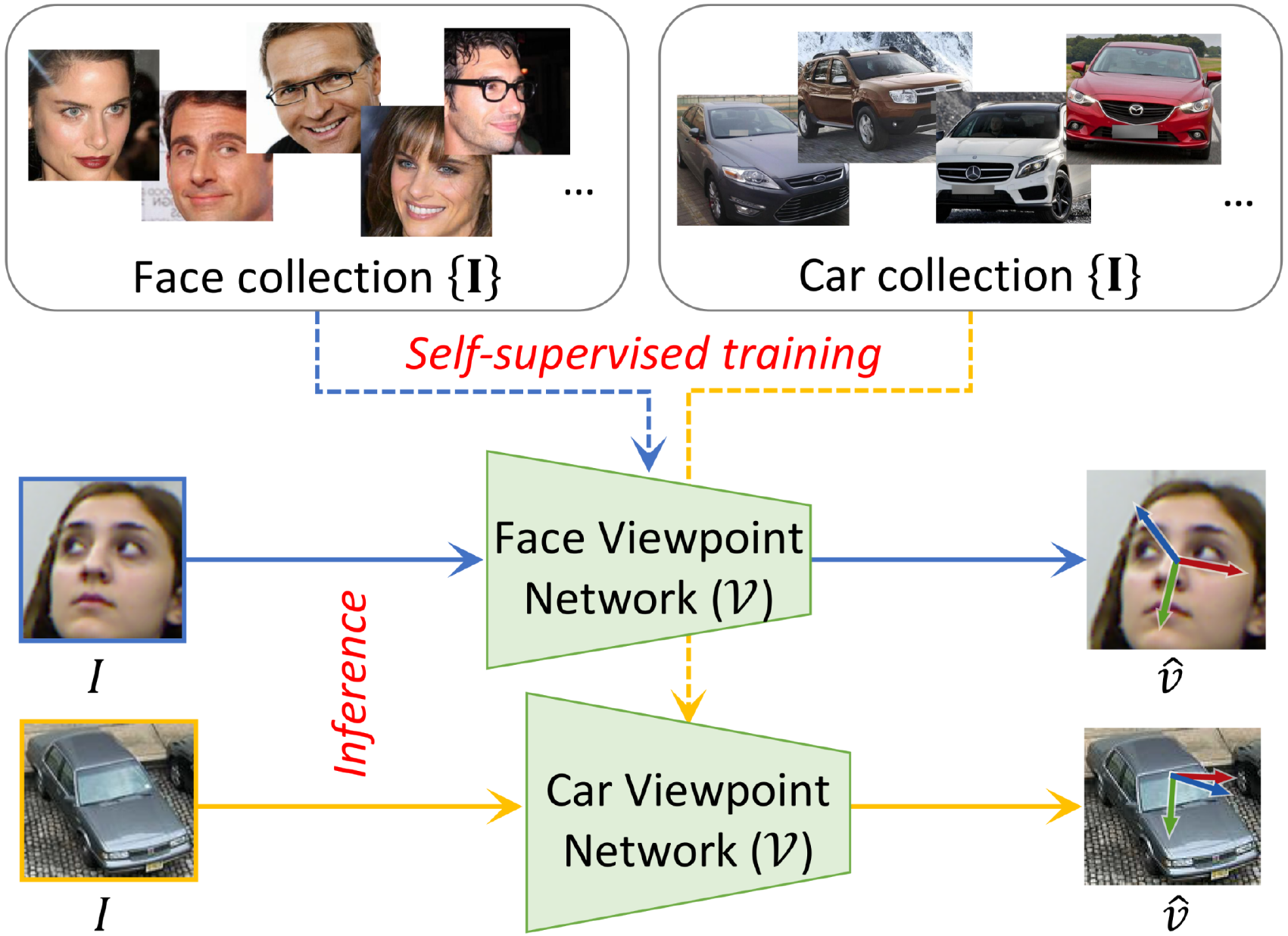}
	\mycaption{Self-supervised viewpoint learning}{We learn a single-image object viewpoint estimation network for each category (face or car) using only a collection of images without ground truth. \vspace{-5mm}}
	\label{fig:teaser}
\end{figure}

In this work, we propose a self-supervised learning technique for viewpoint estimation of general objects that learns from an object image collection without the need for any viewpoint annotations (Figure~\ref{fig:teaser}). By image collection, we mean a set of images containing objects of a category of interest (say, faces or cars). Since viewpoint estimation assumes known object bounding boxes, we also assume that the image collection consists of tightly bounded object images. Being self-supervised in nature, our approach provides an important advancement in viewpoint estimation as it alleviates the need for costly viewpoint annotations. It also enables viewpoint learning on object categories that do not have any existing ground-truth annotations. 

Following the analysis-by-synthesis paradigm, we leverage a viewpoint aware image synthesis network as a form of self-supervision to train our viewpoint estimation network. We couple the viewpoint network with the synthesis network to form a complete cycle and train both together. To self-supervise viewpoint estimation, we leverage cycle-consistency losses between the viewpoint estimation (analysis) network and a viewpoint aware generative (synthesis) network, along with losses for viewpoint and appearance disentanglement, and object-specific symmetry priors. During inference, we only need the viewpoint estimation network, without the synthesis network, making viewpoint inference simple and fast for practical purposes. As per our knowledge, ours is the first self-supervised viewpoint learning framework that learns 3D viewpoint of general objects from image collections in-the-wild. We empirically validate our approach on the human head pose estimation task, which on its own has attracted considerable attention~\cite{zhu2017face, bulat2017far, sun2013deep, yang2018ssr, kumar2017kepler, chang2017faceposenet, gu2017dynamic, yang2019fsa} in computer vision research. We demonstrate that the results obtained by our self-supervised technique are comparable to those of fully-supervised approaches. In addition, we also demonstrate significant performance improvements when compared to viewpoints estimated with self-supervisedly learned keypoint predictors. To showcase the generalization of our technique, we analyzed our approach on object classes such as cars, buses, and trains from the challenging Pascal3D+~\cite{xiang2014beyond} dataset.
We believe this work opens up further research in self-supervised viewpoint learning and would also serve as a robust baseline for future work. 

To summarize, our main contributions are:
\begin{itemize}
\itemsep0em
\item We propose a novel analysis-by-synthesis framework for learning viewpoint estimation in a purely self-supervised manner by leveraging cycle-consistency losses between a viewpoint estimation and a viewpoint aware synthesis network. To our understanding, this is one of first works to explore the problem of self-supervised viewpoint learning for general objects.
\item We introduce generative, symmetric and adversarial constraints which self-supervise viewpoint estimation learning just from object image collections.
\item We perform experiments for head pose estimation on the BIWI dataset \cite{fanelli2013random} and for viewpoint estimation of cars, buses and trains on the challenging Pascal3D+ \cite{xiang2014beyond} dataset and demonstrate competitive accuracy in comparison to fully-supervised approaches.
\end{itemize}

%% file: 02_relatedwork.tex
\section{Related Work}
\vspace{-1mm}
\paragraph{Viewpoint estimation} 

Several successful learning-based viewpoint estimation techniques have been developed for general object categories that either regress orientation directly~\cite{mousavian20173d, mahendran20173d, su2015render, tulsiani2015viewpoints, LiaoCVPR19, prokudin2018deep}; locate 2D keypoints and fit them to 3D keypoints~\cite{grabner20183d, pavlakos20176, zhou2018starmap}; or predict 3D shape and viewpoint parameters~\cite{kundu20183d}. These techniques require object viewpoint annotations during training, either in the form of angular values; or 2D and 3D keypoints and use large annotated datasets, \textit{e.g.}, Pascal3D+~\cite{xiang2014beyond} and ObjectNet3D~\cite{xiang2016objectnet3d} with 12 and 100 categories, respectively. These datasets were annotated via a tedious manual process of aligning best-matched 3D models to images -- a procedure that is not scalable easily to larger numbers of images or categories. To circumvent this problem, existing viewpoint algorithms augment real-world data with synthetic images~\cite{grabner20183d, massaBMVC2016, su2015render, LiaoCVPR19}; assume auxiliary supervision and learn the related aspects (\textit{e.g.}, 3D keypoints) along with viewpoint~\cite{zhou2018starmap, suwajanakorn2018key-pointnet}; or try to learn from very few labeled examples of novel categories~\cite{tseng2019}. 

\vspace{-3mm}
\paragraph{Head pose estimation} Separate from the above-mentioned works, learning-based head pose estimation techniques have also been studied extensively~\cite{zhu2017face, bulat2017far, sun2013deep, yang2018ssr, kumar2017kepler, chang2017faceposenet, gu2017dynamic, yang2019fsa}. These works learn to either predict facial landmarks from data with varying levels of supervision ranging from full~\cite{zhu2017face, bulat2017far, sun2013deep, yang2018ssr, kumar2017kepler}, partial~\cite{honari2018improving}, or no supervision~\cite{hung2019scops, zhang2018unsupervised}; or learn to regress head orientation directly in a fully-supervised manner~\cite{chang2017faceposenet, ruiz2018fine, gu2017dynamic, yang2019fsa}. The latter methods perform better than those that predict facial points~\cite{yang2019fsa}. To avoid manual annotation of head pose, prior works also use synthetic datasets~\cite{zhu2017face, gu2017dynamic}. On the other hand, 
several works~\cite{tewari2018self, feng2018joint, tran2018nonlinear, sahasrabudhe2019lifting} propose
learning-based approaches for dense 3D reconstruction of faces via in-the-wild image collections and some use analysis-by-synthesis~\cite{tewari2018self, tran2018nonlinear}. However, they are not purely self-supervised and use either facial landmarks~\cite{tewari2018self}, dense 3D surfaces~\cite{feng2018joint} or both~\cite{tran2018nonlinear} as supervision.

\vspace{-3mm}
\paragraph{Self-supervised object attribute discovery}
Several recent works try to discover 2D object attributes like landmarks~\cite{zhang2018unsupervised,thewlisICCVT2017, tomas2018neurips} and part segmentation~\cite{hung2019scops,collins2018deep} in a self-supervised manner.
These works are orthogonal to ours as we estimate 3D viewpoint.
Some other works such as~\cite{l2019differ,insafutdinov2018unsupervised,Henderson2018LearningTG} make use of differentiable rendering frameworks to learn 3D shape and/or camera viewpoint from a single or multi-view image collections.
Because of heavy reliance on differentiable rendering, these works
mainly operate on synthetic images. In contrast, our approach can learn viewpoints from image collections in the wild. Some works learn 3D reconstruction from in-the-wild image collections, but use annotated object silhouettes along with other annotations such as 2D semantic keypoints~\cite{cmrKanazawa18}, category-level 3D templates~\cite{kulkarni2019csm}; or multiple views of each object instance~\cite{kato2018renderer, Wiles17a, novotny2017learning}. In contrast, we use no additional supervision other than the image collections that comprise of independent object images. To the best we know, no prior works propose to learn viewpoint of general objects in a purely self-supervised manner from in-the-wild image collections.

%% file: 03_method.tex
\section{Self-Supervised Viewpoint Learning}

\noindent \textbf{Problem setup}
We learn a viewpoint estimation network~$\mathcal{V}$ using an in-the-wild image collection $\{\textbf{I}\}$ of a specific object category without annotations. Since viewpoint estimation assumes tightly cropped object images, we also assume that our image collection is composed of cropped object images. Figure~\ref{fig:teaser} shows some samples in the face and car image collections. During inference, the viewpoint network $\mathcal{V}$ takes a single object image \textit{I} as input and predicts the object 3D viewpoint $\boldsymbol{\hat{v}}$.

\vspace{1mm}
\noindent \textbf{Viewpoint representation} To represent an object viewpoint $\boldsymbol{\hat{v}}$, we use three Euler angles, namely azimuth ($\hat{a}$), elevation ($\hat{e}$) and in-plane rotation ($\hat{t}$) describing the rotations around fixed 3D axes.
For the ease of viewpoint regression, we represent each Euler angle, \eg, $a \in [0,2\pi]$, as a point on a unit circle with 2D coordinates $(\cos(a),\sin(a))$.
Following~\cite{LiaoCVPR19}, instead of predicting co-ordinates on a 360$^\circ$ circle, we predict a positive unit vector in the first quadrant with $|\hat{a}|$ = $(|\cos(\hat{a})|,|\sin(\hat{a})|)$ and also the category of the combination of signs of $\sin(\hat{a})$ and $\cos(\hat{a})$ indicated by $\sign(\hat{a})$ = $(\sign(\cos(\hat{a})), \sign(\sin(\hat{a})) \in \{(+,+), (+,-), (-,+), (-,-)\}$.
Given the predicted $|\hat{a}|$ and $\sign(\hat{a})$ from the viewpoint network, we can construct $cos(\hat{a})$ = $\sign(\cos(\hat{a})) |\cos(\hat{a})|$ and $\sin(\hat{a})$ = $\sign(\sin(\hat{a})) |\sin(\hat{a})|$. The predicted Euler angle $\hat{a}$ can finally be computed as  $\tanh(\sin(\hat{a})/\cos(\hat{a}))$. In short, the viewpoint network performs both regression to predict a positive unit vector $|a|$ and also classification to predict the probability of $\sign(a)$.

\begin{figure}
	\centering
	\includegraphics[width=\linewidth]{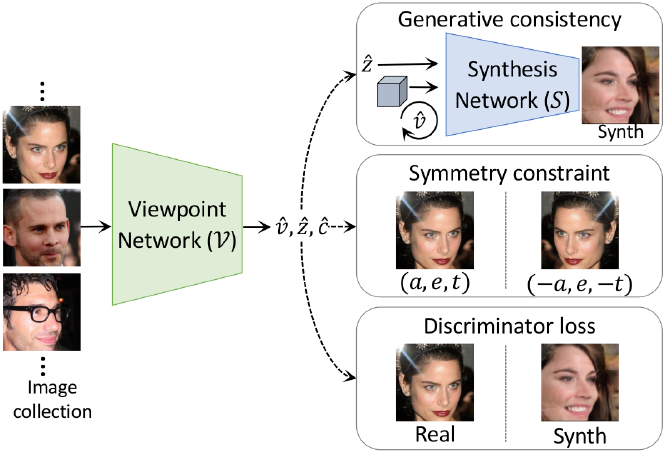}
	\mycaption{Approach overview}{We use generative consistency, symmetry and discriminator losses to supervise the viewpoint network with a collection of images without annotations. \vspace{-5mm} }
	\label{fig:vpnet_losses}
\end{figure}

\vspace{1mm}
\noindent \textbf{Approach overview and motivation}
We learn the viewpoint network $\mathcal{V}$ using a set of self-supervised losses as illustrated in Figure~\ref{fig:vpnet_losses}. To formulate these losses we use three different constraints, namely generative consistency, a symmetry constraint and a discriminator loss.
Generative consistency forms the core of the self-supervised constraints to train our viewpoint network and is inspired from the popular analysis-by-synthesis learning paradigm~\cite{kundu20183d}. This framework tries to tackle inverse problems (such as viewpoint estimation) by modelling the forward process of image or feature synthesis.  A synthesis function $\mathcal{S}$ models the process of generating an image of an object from a basic representation and a set of parameters. The goal of the analysis function is to infer the underlying parameters which can best explain the formation of an observed input image. Bayesian frameworks such as~\cite{yuille2006vision} and inverse graphics~\cite{kundu20183d, kato2018renderer, yao20183d, liu2019softras,jampani2015informed} form some of the popular techniques that are based on the analysis-by-synthesis paradigm. In our setup, we consider the viewpoint network $\mathcal{V}$ as the analysis function.

\begin{figure}
	\centering
	\includegraphics[width=1\linewidth]{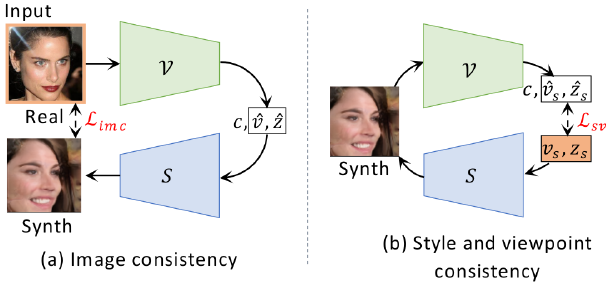}
	\mycaption{Generative consistency}{The two cyclic (a) image consistency ($\mathcal{L}_\mathit{imc}$) and (b) style and viewpoint consistency  ($\mathcal{L}_\mathit{sv}$) losses make up generative consistency. The input to each cycle is highlighted in yellow. Image consistency enforces that an input real image, after viewpoint estimation and synthesis, matches its reconstructed synthetic version. Style and viewpoint consistency enforces that the input style and viewpoint provided for synthesis are correctly reproduced by the viewpoint network.
	\vspace{-5mm}}
	\label{fig:gencon_cycles}
\end{figure}

We model the synthesis function $\mathcal{S}$, with a viewpoint aware image generation model. 
Recent advances in Generative Adversarial Networks (GAN) \cite{chen2016infogan,karras2019style,nguyenhologan} have shown that it is possible to generate high-quality images with fine-grained control over parameters like appearance, style, viewpoint, etc. Inspired by these works, our synthesis network generates an image, given an input  $\boldsymbol{{v}}$, which controls the viewpoint of the object and an input vector $\boldsymbol{{z}}$, which controls the style of the object in the synthesized image.
By coupling both the analysis ($\mathcal{V}$) and synthesis ($\mathcal{S}$) networks in a cycle, we learn both the networks in a self-supervised manner using cyclic consistency constraints described in \ref{sec::vpnet_gencon_loss} and shown in Figure~\ref{fig:gencon_cycles}. Since the synthesis network can generate high quality images based on controllable inputs $\boldsymbol{{v}}$ and $\boldsymbol{{z}}$, these synthesized images can in turn be used as input to the analysis network ($\mathcal{V}$) along with $\boldsymbol{{v}}$, $\boldsymbol{{z}}$ as the pseudo ground-truth. On the other hand, for a real world image, if $\mathcal{V}$ predicts the correct viewpoint and style, these can be utilized by $\mathcal{S}$ to produce a similar looking image. This effectively functions as image reconstruction-based supervision. In addition to this, similar to~\cite{chen2016infogan,nguyenhologan} the analysis network also functions as a discriminator, evaluating whether the synthesized images are real or fake. Using a widely prevalent observation that several real-world objects are symmetric, we also enforce a prior constraint via a symmetry loss function to train the viewpoint network. 
Object symmetry has been used in previous supervised techniques such as~\cite{mahendran20173d} for data augmentation, but not as a loss function. 
In the following, we first describe the various loss constraints used to train the viewpoint network $\mathcal{V}$ while assuming that we already have a trained synthesis network $\mathcal{S}$. In Section~\ref{sec::vgs}, we describe the loss constraints used to train the synthesis network $\mathcal{S}$.

\subsection{Generative Consistency} 
\label{sec::vpnet_gencon_loss}
\vspace{-1mm}

As Figure~\ref{fig:gencon_cycles} illustrates, we couple the viewpoint network $\mathcal{V}$ with the synthesis network $\mathcal{S}$ to create a circular flow of information resulting in two consistency losses: (a) image consistency and (b) style and viewpoint consistency.

\vspace{1mm}
\noindent \textbf{Image consistency}
Given a real image $I$ sampled from a given image collection $\{\textbf{I}\}$, we first predict its viewpoint $\boldsymbol{\hat{v}}$ and style code $\boldsymbol{\hat{z}}$ via the viewpoint network $\mathcal{V}$. Then, we pass the predicted $\boldsymbol{\hat{v}}$ and $\boldsymbol{\hat{z}}$ into the synthesis network $\mathcal{S}$ to create the synthetic image $\hat{I}_s$. To train the viewpoint network, we use the image consistency between the input image $I$ and corresponding synthetic image $I_s$ with a perceptual loss:
\vspace{-1mm}
\begin{equation}
\mathcal{L}_\mathit{imc} = 1 - \langle \Phi(I),\Phi(\hat{I}_s) \rangle, 
\vspace{-1mm}
\end{equation}
where $\Phi(.)$ denotes the conv5 features of an ImageNet-trained~\cite{deng2009imagenet} VGG16 classifier~\cite{Simonyan15_vgg} and $\langle .,.\rangle$ denotes the cosine distance.
Figure~\ref{fig:gencon_cycles}(a) illustrates the image consistency cycle.

\vspace{1mm}
\noindent \textbf{Style and viewpoint consistency} 
As illustrated in Figure~\ref{fig:gencon_cycles}(b), we create another circular flow of information with the viewpoint and synthesis networks, but this time starting with a random
viewpoint $\boldsymbol{{v}}_s$ and a style code $\boldsymbol{{z}}_s$, both sampled from uniform distributions, and input them to the synthesis network to create an image $I_s = \mathcal{S}(\boldsymbol{v}_s, \boldsymbol{z}_s)$. We then pass the synthetic image $I_s$ to the viewpoint network $\mathcal{V}$ that predicts its viewpoint $\boldsymbol{\hat{v}}_s$ and the style code $\boldsymbol{\hat{z}}_s$.
We use the sampled viewpoint and style codes for the synthetic image $I_s$ as a pseudo GT to train the viewpoint network. Following \cite{LiaoCVPR19}, the viewpoint consistency loss $\mathcal{L}_v(\boldsymbol{\hat{v}_1}, \boldsymbol{\hat{v}_2})$ between two viewpoints $\boldsymbol{\hat{v}_1}=(\hat{a}_1,\hat{e}_1,\hat{t}_1)$ and $\boldsymbol{\hat{v}_2}=(\hat{a}_2,\hat{e}_2,\hat{t}_2)$ has two components for each Euler angle: (i) cosine proximity between the positive unit vectors $\mathcal{L}_{v}^{|a|} = -\langle |\hat{a}_1)|, |\hat{a}_2|\rangle$ and (ii) the cross-entropy loss $\mathcal{L}_{v}^{\sign(a)}$ between the classification probabilities of $\sign(\hat{a}_1)$ and $\sign(\hat{a}_2)$. The viewpoint consistency loss $\mathcal{L}_v$ is a sum of the cross-entropy and cosine proximity losses for all the three Euler angles:
\vspace{-2mm}
\begin{equation}
\mathcal{L}_v(\boldsymbol{\hat{v}_1}, \boldsymbol{\hat{v}_2}) = \sum_{\phi  \in  {a,e,t}} \mathcal{L}_v^{|\phi|} +  \mathcal{L}_v^{\sign(\phi)}.
\label{eq:viewpoint_loss}
\vspace{-3mm}
\end{equation}
The overall style and viewpoint loss between the sampled $(\boldsymbol{v}_s, \boldsymbol{z}_s)$ and the predicted $(\boldsymbol{\hat{v}}_s, \boldsymbol{\hat{z}}_s)$ is hence:
\vspace{-1mm}
\begin{equation}
\mathcal{L}_{sv} = \left\lVert \boldsymbol{z}_s- \boldsymbol{\hat{z}}_s\right\rVert^{2}_{2} + \mathcal{L}_v(\boldsymbol{v}_s, \boldsymbol{\hat{v}}_s).
\label{eq:style_view_loss}
\vspace{-1mm}
\end{equation}
While viewpoint consistency enforces that $\mathcal{V}$ learns correct viewpoints for synthetic images, image consistency helps to ensure that $\mathcal{V}$ generalizes well to real images as well, and hence avoids over-fitting to images synthesized by $\mathcal{S}$.

\subsection{Discriminator Loss} 
\label{sec::vpnet_discr_loss}
\vspace{-2mm}
$\mathcal{V}$ also predicts a score $\boldsymbol{\hat{c}}$ indicating whether an input image is real or synthetic. It thus acts as a discriminator in a typical GAN~\cite{goodfellow2014generative} setting, helping the synthesis network create more realistic images. 
We use the discriminator loss from Wasserstein-GAN~\cite{arjovsky2017wasserstein} to update the viewpoint network using:
\vspace{-3mm}
\begin{equation}
\mathcal{L}_\mathit{dis} = -\mathbb{E}_{x \sim p_\mathrm{real}}[\boldsymbol{c}] + \mathbb{E}_{\hat{x} \sim p_\mathrm{synth}}[\hat{\boldsymbol{c}}],
\label{eqn:discr_loss}
\vspace{-1mm}
\end{equation}
where $\boldsymbol{c}$ = $\mathcal{V}(x)$ and $\boldsymbol{\hat{c}}$ = $\mathcal{V}(\hat{x})$ are the predicted class scores for the real and the synthesized images, respectively.

\subsection{Symmetry Constraint}
\label{sec::vpnet_symloss} 
\vspace{-1mm}
Symmetry is a strong prior observed in many commonplace object categories, \textit{e.g.}, faces, boats, cars, airplanes, etc. For categories with symmetry, we propose to leverage an additional symmetry constraint. Given an image \textit{I} of an object with viewpoint $(a, e, t)$, the GT viewpoint of the object in a horizontally flipped image $flip(I)$ is given by $($-$a, e, $-$t)$.
We enforce a symmetry constraint on the viewpoint network's outputs $(\boldsymbol{\hat{v}}, \boldsymbol{\hat{z}})$ and $(\boldsymbol{\hat{v}}^*,\boldsymbol{\hat{z}}^*)$ for a given image \textit{I} and its horizontally flipped version $flip(I)$, respectively.
Let $\boldsymbol{\hat{v}}$=$(\hat{a}, \hat{e}, \hat{t})$ and $\boldsymbol{\hat{v}}^*$=$(\hat{a}^*, \hat{e}^*, \hat{t}^*)$
and we denote the flipped viewpoint of the flipped image as  $\boldsymbol{\hat{v}}_f^*$=$($-$\hat{a}^*, \hat{e}^*, $-$\hat{t}^*)$.
The symmetry loss is given as
\vspace{-2mm}
\begin{equation}
\mathcal{L}_\mathit{sym} = \mathcal{D}(\boldsymbol{\hat{v}}, \boldsymbol{\hat{v}}_f^*) + 
\left\lVert \boldsymbol{{\hat{z}}}- \boldsymbol{\hat{z}^*}\right\rVert^{2}_{2}.
\vspace{-2mm}
\end{equation}
Effectively, for a given horizontally flipped image pair, we regularize that the network predicts similar magnitudes for all the angles and opposite directions for azimuth and tilt. Additionally, the above loss enforces that the style of the flipped image pair is consistent.

Our overall loss to train the viewpoint network $\mathcal{V}$ is a linear combination of the aforementioned loss functions:
\vspace{-2mm}
\begin{equation}
\mathcal{L}_{V} = \lambda_1\mathcal{L}_\mathit{sym} + \lambda_2\mathcal{L}_\mathit{imc} + \lambda_3\mathcal{L}_{sv} + \lambda_4\mathcal{L}_\mathit{dis},
\vspace{-2mm}
\end{equation}

\noindent where the parameters $\{\lambda_i\}$ determine the relative importance of the different losses, which we empirically determine using a grid search.
\begin{figure}
	\centering
	\includegraphics[width=\linewidth]{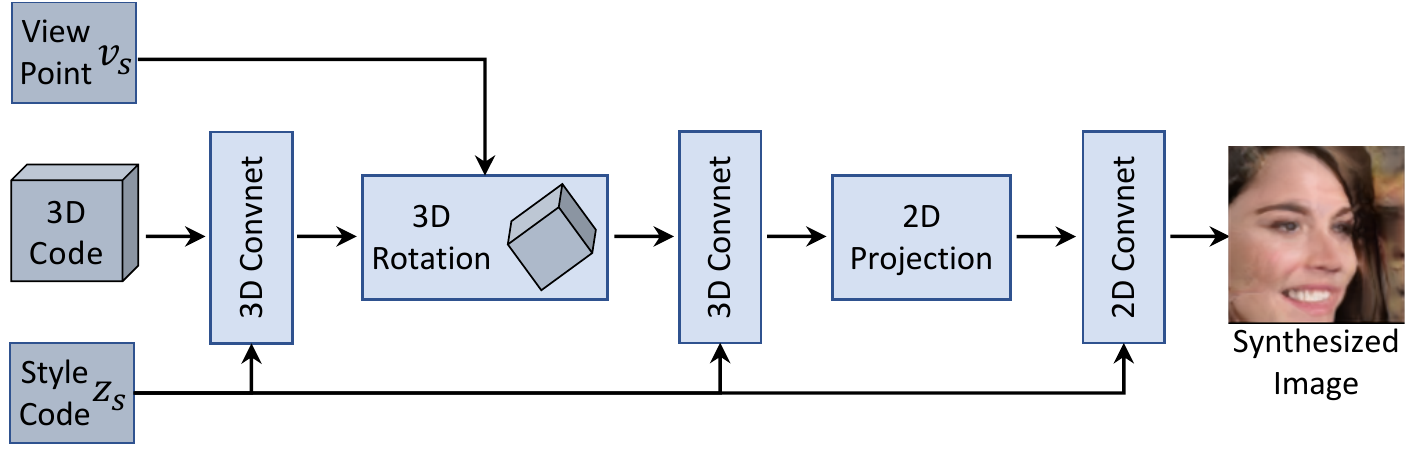}
	\mycaption{Synthesis network overview}{The network takes viewpoint $\boldsymbol{v_s}$ and style code $\boldsymbol{z_s}$ to produce a viewpoint aware image. \vspace{-7mm}}
	\label{fig:synthesis_net}
\end{figure}

\vspace{-2mm}
\section{Viewpoint-Aware Synthesis Network}
\label{sec::vgs}
\vspace{-1mm}

Recent advances in GANs such as InfoGAN~\cite{chen2016infogan},  StyleGAN~\cite{karras2019style} and HoloGAN~\cite{nguyenhologan} demonstrate the
possibility of conditional image synthesis where we can control the synthesized object's attributes such as object class, 
viewpoint, style, geometry, etc.
A key insight that we make use of in our synthesis network and which is also used in recent GANs such as HoloGAN~\cite{nguyenhologan} and other works\cite{zhu2018visual,kulkarni2015deep,sitzmann2019deepvoxels}, is that one can instill 3D geometric meaning into the network's latent representations by performing explicit geometric transformations such as rotation on them. A similar idea has also been used successfully with other generative models such as auto-encoders~\cite{Hinton2011ICANN, Rhodin2018ECCV,Park2019ICCV}.
Our viewpoint-aware synthesis network has a similar architecture to HoloGAN~\cite{nguyenhologan}, but is tailored for the needs of 
viewpoint estimation. HoloGAN is a pure generative model with GAN losses to ensure realism and an identity loss to reproduce the input style code, but lacks a corresponding viewpoint prediction network. 
In this work, since we focus on viewpoint estimation, we introduce tight coupling of HoloGAN with a viewpoint prediction network and several novel loss functions to train it in a manner that is conducive to accurate viewpoint prediction.

\vspace{1mm}
\noindent \textbf{Synthesis network overview} Figure~\ref{fig:synthesis_net} illustrates the design of the synthesis network.
The network $\mathcal{S}$ takes a style code $\boldsymbol{z}_s$ and a viewpoint $\boldsymbol{v}_s$ to produce a corresponding object image $I_s$. The goal of $\mathcal{S}$ is to learn a disentangled 3D representation of an object, which can be used to synthesize objects in various viewpoints and styles, hence aiding in the supervision of the viewpoint network $\mathcal{V}$.
We first pass a learnable canonical 3D latent code through a 3D network, which applies 3D convolutions to it.
Then, we rotate the resulting 3D representation with $\boldsymbol{v}_s$ and pass through an additional 3D network.
We project this viewpoint-aware learned 3D code on to 2D using a simple orthographic projection unit. Finally, we pass the resulting 2D representation through a StyleGAN~\cite{karras2019style}-like 2D network to produce a synthesized image. The style and appearance of the image is
controlled by the sampled style code $\boldsymbol{z}_s$. Following StyleGAN~\cite{karras2019style}, the style code $\boldsymbol{z}_s$ affects the style of the resulting image via adaptive instance normalization~\cite{huang2017arbitrary} in both the 3D and 2D representations.
For stable training, we freeze $\mathcal{V}$ while training $\mathcal{S}$ and vice versa.

\vspace{1mm}
\noindent \textbf{Loss functions}
Like the viewpoint network, we use several constraints to train the synthesis network, which are designed to improve viewpoint estimation. The first is the standard adversarial loss used in training Wasserstein-GAN\cite{arjovsky2017wasserstein}: 
\vspace{-1mm}
\begin{equation}
\mathcal{L}_\mathit{adv} = -\mathbb{E}_{\hat{x} \sim p_\mathrm{synth}}[\boldsymbol{\hat{c}}]
\label{eqn:adv_loss}
\vspace{-1mm}
\end{equation}
where $\boldsymbol{\hat{c}}$ = $\mathcal{V}(\hat{x})$ is the class membership score predicted by $\mathcal{V}$ for a synthesized image. The second is a paired version of the style and viewpoint consistency loss (Eqn.~\ref{eq:style_view_loss}) described in Section~\ref{sec::vpnet_gencon_loss}, where we propose to use multiple paired $(\boldsymbol{z}_s,\boldsymbol{v}_s)$ samples to enforce style and viewpoint consistency and to better disentangle the latent representations of $\mathcal{S}$. The third is a flip image consistency loss. Note that, in contrast to our work, InfoGAN~\cite{chen2016infogan} and HoloGAN~\cite{nguyenhologan} only use adversarial and style consistency losses.

\begin{figure}[t!]
	\centering
	\includegraphics[width=\linewidth]{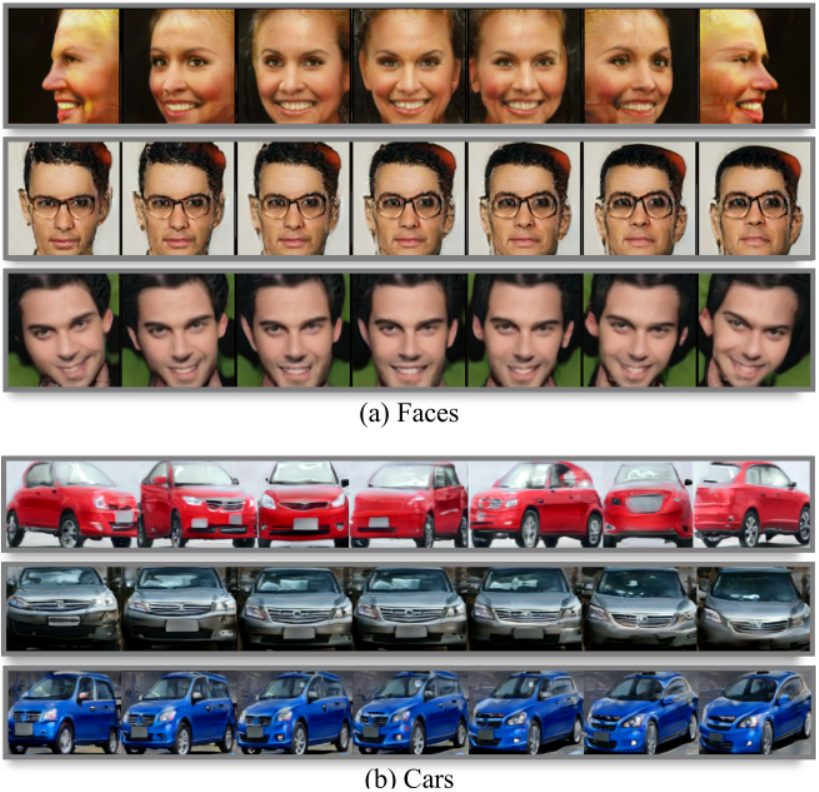}
	\mycaption{Synthesis results}{Example synthetic images of (a) faces and (b) cars generated by the viewpoint-aware generator $\mathcal{S}$. For each row the style vector $z$ is constant, whereas the viewpoint is varied monotonically along the azimuth (first row), elevation (second row) and tilt (third row) dimensions.  
	\vspace{-5mm}}
	\label{fig:sythesis}
\end{figure}

\noindent \textbf{Style and viewpoint consistency with paired samples} 
 Since we train the viewpoint network with images synthesized by $\mathcal{S}$, it is very important for $\mathcal{S}$ to be sensitive and responsive to its input style $\boldsymbol{z}_s$ and viewpoint $\boldsymbol{v}_s$ parameters. An ideal $\mathcal{S}$ would perfectly disentangle $\boldsymbol{v}_s$ and $\boldsymbol{z}_s$. That means, if we fix $\boldsymbol{z}_s$ and vary $\boldsymbol{v}_s$, the resulting object images should have the same style, but varying viewpoints. On the other hand, if we fix $\boldsymbol{v}_s$ and vary $\boldsymbol{z}_s$, the resulting object images should have different styles, but a fixed viewpoint. We enforce this constraint with a paired version of the style and viewpoint consistency (Eqn.~\ref{eq:style_view_loss}) loss where we sample 3 different pairs of $(\boldsymbol{z}_s,\boldsymbol{v}_s)$ values by varying one parameter at a time as: $\{(\boldsymbol{z}_s^0,\boldsymbol{v}_s^0),(\boldsymbol{z}_s^0,\boldsymbol{v}_s^1),(\boldsymbol{z}_s^1,\boldsymbol{v}_s^1)\}$. We refer to this paired style and viewpoint loss as $\mathcal{L}_\mathit{sv, pair}$. The ablation study in Section~\ref{sec:experiments} suggests that this paired style and viewpoint loss helps to train a better synthesis network for our intended task of viewpoint estimation. We also observe qualitatively that the synthesis network successfully disentangles the viewpoints and styles of the generated images. Some example images synthesized by $\mathcal{S}$ for faces and cars are shown in Figure~\ref{fig:sythesis}. Each row uses a fixed style code $\boldsymbol{z}_s$ and we monotonically vary the input viewpoint $\boldsymbol{v}_s$ by changing one of its $a$, $e$ or $t$ values across the columns.

\vspace{1mm}
\noindent \textbf{Flip image consistency}
This is similar to the symmetry constraint used to train the viewpoint network, but applied to synthesized images.
Flip image consistency forces $\mathcal{S}$ to synthesize horizontally flipped images when we input appropriately flipped viewpoints. 
For the pairs $\mathcal{S}(\boldsymbol{v}_s,\boldsymbol{z}_s) = I_s$ and $\mathcal{S}(\boldsymbol{v}_s^*,\boldsymbol{z}_s) = I_s^*$, where $\boldsymbol{v}^*$ has opposite signs for the $a$ and $t$ values of $\boldsymbol{v}_s$, the flip consistency loss is defined as:
\vspace{-2mm}
\begin{equation}
\mathcal{L}_{fc} = \left\lVert I_s - \mathrm{flip}(I_s^*) \right\rVert_{1}
\vspace{-2mm}
\end{equation}
where $\mathrm{flip}(I_s^*)$ is the horizontally flipped version of $I_s^*$.

The overall loss for the synthesis network is given by:
\vspace{-2mm}
\begin{equation}
\mathcal{L}_{S} = \lambda_5\mathcal{L}_\mathit{adv} + \lambda_6\mathcal{L}_{sv, pair} + \lambda_7\mathcal{L}_{fc}
\vspace{-2mm}
\end{equation}
where the parameters $\{\lambda_i\}$ are the relative weights of the losses which we determine empirically using grid search.

%% file: 04_results.tex
\vspace{-1mm}
\section{Experiments}
\label{sec:experiments}

\vspace{-1mm}
We empirically validate our approach with extensive experiments on head pose estimation and viewpoint estimation on other object categories of buses, cars and trains. We refer to our approach as `SSV'.

\vspace{1mm}
\noindent \textbf{Implementation and training details}
We implement our framework in Pytorch\cite{paszke2017automatic}. We provide all network architecture details, and run-time and memory analyses in the supplementary material. 

\vspace{1mm}
\noindent \textbf{Viewpoint calibration} The output of SSV for a given image $I$ is ($\hat{a}, \hat{e}, \hat{t}$). However, since SSV is self-supervised, the co-ordinate system for predictions need
not correspond to the actual canonical co-ordinate system of GT annotations.
For quantitative evaluation, following the standard practice in self-supervised learning of features~\cite{donahue2016adversarial, zhang2017split, caron2018deep}
and landmarks~\cite{hung2019scops, zhang2018unsupervised,thewlisICCVT2017}, we fit a linear regressor that maps the predictions of SSV to GT viewpoints using 100 randomly chosen images from the target test dataset. 
Note that this calibration with a linear regressor only rotates the predicted viewpoints to the GT canonical frame of reference. We do not update or learn our SSV network during this step.  

\subsection{Head Pose Estimation}
\vspace{-2mm}
Human faces have a special place among objects for viewpoint estimation and head pose estimation has attracted considerable research attention~\cite{zhu2017face, bulat2017far, sun2013deep, yang2018ssr, kumar2017kepler, chang2017faceposenet, gu2017dynamic, yang2019fsa}. The availability of large-scale datasets~\cite{300wlp, fanelli2013random} and the existence of ample research provides a unique opportunity to perform extensive experimental analysis
of our technique on head pose estimation.

\vspace{1mm}
\noindent \textbf{Datasets and evaluation metric}
For training, we use the 300W-LP~\cite{300wlp} dataset, which combines several in-the-wild face datasets. It contains 122,450 face images with diverse viewpoints, created by fitting a 3D face morphable model~\cite{blanz1999morphable} to face images and rendering them from various viewpoints.
Note that we only use the images from this dataset to train SSV and not their GT viewpoint annotations. 
We evaluate our framework on the 
BIWI~\cite{fanelli2013random} dataset which contains 15,677 images across 24 sets of video sequences of 20 subjects in a wide variety of viewpoints. We use the MTCNN face detector to detect all faces~\cite{zhang2016joint}. 
We compute average absolute errors (AE) for azimuth, elevation and tilt between the predictions and GT. We also report the mean absolute error (MAE) of these three errors.

\setlength{\tabcolsep}{3pt} 
\begin{table}\centering
\begin{tabular}{clcccc}
\toprule
& \multicolumn{1}{l}{Method} & \multicolumn{1}{c}{Azimuth} & \multicolumn{1}{c}{Elevation} & \multicolumn{1}{c}{Tilt} & \multicolumn{1}{c}{MAE}\\
\midrule
\small
\parbox[t]{3mm}{\multirow{8}{*}{\rotatebox[origin=c]{90}{Self-Supervised}}}
\scriptsize
    & LMDIS~\cite{zhang2018unsupervised} + PnP                 & 16.8 & 26.1 & 5.6  & 16.1 \\
    & IMM~\cite{tomas2018neurips} + PnP                 & 14.8 & 22.4 & 5.5  & 14.2 \\
    & SCOPS~\cite{hung2019scops} + PnP                       & 15.7 & 13.8 & 7.3  & 12.3 \\
	& HoloGAN~\cite{nguyenhologan}                                      &   8.9  & 15.5 & 5.0 & 9.8 \\
    & HoloGAN~\cite{nguyenhologan} with $v$                            & 7.0 & 15.1 & 5.1  & 9.0    \\
    \cdashline{2-6}
    \rule{0pt}{3ex}
    & SSV w/o $\mathcal{L}_\mathit{sym}$ + $\mathcal{L}_\mathit{imc}$                                      & 6.8    & 13.0   & 5.2    & 8.3    \\
    & SSV w/o $\mathcal{L}_\mathit{imc}$                         & 6.9  & 10.3 & \textbf{4.4}  & 7.2  \\
    & SSV-Full  & \textbf{6.0}  & \textbf{9.8}  & \textbf{4.4}  & \textbf{6.7}  \\
    \midrule
\parbox[t]{3mm}{\multirow{6}{*}{\rotatebox[origin=c]{90}{Supervised}}}    
    & 3DDFA~\cite{zhu2017face}                       & 36.2 & 12.3 & 8.7  & 19.1 \\
    & KEPLER~\cite{kumar2017kepler}                        & 8.8  & 17.3 & 16.2 & 13.9 \\
    & DLib~\cite{kazemi2014one}                       & 16.8 & 13.8 & 6.1  & 12.2 \\
    & FAN~\cite{bulat2017far}          & 8.5  & 7.4  & 7.6  & 7.8  \\
    & Hopenet~\cite{ruiz2018fine}      & 5.1  & 6.9  & 3.3  & 5.1  \\
    & FSA~\cite{yang2019fsa}                               & \textbf{4.2}  & \textbf{4.9}  & \textbf{2.7}  & \textbf{4.0}   \\
\midrule
\end{tabular}
\mycaption{Head pose estimation ablation studies and SOTA comparisons}{Average absolute angular error for azimuth, elevation and tilt Euler angles in degrees together with the mean absolute error (MAE) for the BIWI~\cite{fanelli2013random} dataset.}
\label{tab::faces_ablation}
\vspace{-5mm}
\end{table}

\vspace{1mm}
\noindent\textbf{Ablation study} We empirically evaluate the different self-supervised constraints used to train the viewpoint network.
Table~\ref{tab::faces_ablation} shows that for head pose estimation, using all the proposed constraints (SSV-Full) results in our best MAE of \ang{6.7}. Removing the image consistency constraint $\mathcal{L}_\mathit{imc}$ leads to an MAE to \ang{7.2} and further
removing the symmetry constraint $\mathcal{L}_\mathit{sym}$ results in an MAE of \ang{8.3}.
These results demonstrate the usefulness of the generative image consistency and symmetry constraints in our framework.

\setlength{\tabcolsep}{3pt} 
\begin{table}\centering
\begin{tabular}{llcccc}
\toprule
&\multicolumn{1}{l}{Method} & \multicolumn{1}{c}{Azimuth} & \multicolumn{1}{c}{Elevation} & \multicolumn{1}{c}{Tilt} & \multicolumn{1}{c}{MAE}\\[2pt]
\midrule
\small
\parbox[t]{4mm}{\multirow{3}{*}{\rotatebox[origin=l]{90}{Self-Sup}}}
\scriptsize
\rule{0pt}{1ex}
&SSV non-refined                                 & 6.9    & 9.4  & \textbf{4.2}  & 6.8  \\[2pt]
&SSV refined on BIWI                             & \textbf{4.9}  & \textbf{8.5}   & \textbf{4.2} & \textbf{5.8}   \\[4pt]
\midrule
\small
\parbox[t]{2mm}{\multirow{3}{*}{\rotatebox[origin=l]{90}{Supervised}}}
\scriptsize
&FSA~\cite{yang2019fsa}                           & \textbf{2.8}  & 4.2  & 3.6  & \textbf{3.6}  \\
\addlinespace[1pt]
&DeepHP~\cite{mukherjee2015deep}         & 5.6  & 5.1  &  -   &  -   \\
\addlinespace[1pt]
&RNNFace~\cite{gu2017dynamic}                                 &3.9  & \textbf{4.0}  &  \textbf{3.0}   &  \textbf{3.6}   \\
\addlinespace[2pt]
\midrule
\end{tabular}
\mycaption{Improved head pose estimation with fine-tuning}{Average angular error for each of the Euler angles together with mean average error (MAE) on data of 30\% held-out sequences of the BIWI~\cite{fanelli2013random} dataset and fine-tuning on the remaining 70\% without using their annotations. All values are in degrees.}
\label{tab::biwi_refinement}
\vspace{-6mm}
\end{table}
Additionally, we evaluate the effect of using the paired style and viewpoint loss $\mathcal{L}_\mathit{sv, pair}$ to train the viewpoint-aware synthesis network $\mathcal{S}$. We observe that when we train $\mathcal{S}$ without $\mathcal{L}_\mathit{sv, pair}$, our viewpoint network (SSV-full model) results in AE values of $7.8^{\circ}$ (azimuth), $11.1^{\circ}$ (elevation), $4.2^{\circ}$ (tilt) and an MAE of $7.7^{\circ}$. This represents a $1^{\circ}$ increase from the corresponding MAE value of $6.7^{\circ}$ for the SSV-full, where $\mathcal{S}$ is trained with $\mathcal{L}_\mathit{sv, pair}$ (Table~\ref{tab::faces_ablation}, SSV-full). This shows that our paired style and viewpoint loss helps to better train the image synthesis network for the task of viewpoint estimation.

\vspace{1mm}
\noindent \textbf{Comparison with self-supervised methods}
Since SSV is a self-supervised viewpoint estimation work, there is no existing work that we can directly compare against. One could also obtain head pose from predicted face landmarks and we compare against recent state-of-the-art self-supervised landmark estimation (LMDIS~\cite{zhang2018unsupervised}, IMM~\cite{tomas2018neurips}) and part discovery techniques (SCOPS~\cite{hung2019scops}).
We fit a linear regressor that maps the self-supervisedly learned semantic face part centers from SCOPS and landmarks from LMDIS, IMM to five canonical facial landmarks (left-eye center, right-eye center, nose tip and mouth corners). Then we fit an average 3D face model to these facial landmarks with the Perspective-n-Point (PnP) algorithm~\cite{lepetit2009epnp} to estimate head pose. We also quantify   HoloGAN's~\cite{nguyenhologan} performance at viewpoint estimation, by training a viewpoint network with images synthesized by it under different input viewpoints (as pseudo GT). Alternatively, we train HoloGAN with an additional viewpoint output and a corresponding additional loss for it. For both these latter approaches, we additionally use viewpoint calibration, similar to SSV. We consider these works as our closest baselines because of their self-supervised training. The MAE results in Table~\ref{tab::faces_ablation} indicate that SSV performs considerably better than all the competing self-supervised methods.

\vspace{1mm}
\noindent\textbf{Comparison with supervised methods}
As a reference, we also report the metrics for the recent state-of-the-art fully-supervised methods.
Table~\ref{tab::faces_ablation} shows the results for both
the keypoint-based~\cite{zhu2017face, kumar2017kepler, kazemi2014one, bulat2017far} and keypoint-free~\cite{ruiz2018fine, yang2019fsa} methods. The latter methods learn to directly regress head orientation values from networks. The results indicate that `SSV-Full', despite being purely self-supervised, can obtain
comparable results to fully supervised techniques. In addition, we notice that
SSV-Full (with MAE \ang{6.7}) outperforms all the keypoint-based supervised methods~\cite{zhu2017face, kumar2017kepler, kazemi2014one, bulat2017far}, where FAN~\cite{bulat2017far} has the best MAE of \ang{7.8}.

\vspace{1mm}
\noindent\textbf{Refinement on BIWI dataset}
The results reported thus far are with training on the 300W-LP~\cite{300wlp} dataset.
Following some recent works~\cite{yang2019fsa, mukherjee2015deep, gu2017dynamic}, we use 70\% (16) of the image sequences in the BIWI dataset to fine-tune our model. Since our method is self-supervised, we just use images from BIWI without the annotations. We use the remaining 30\% (8) image sequences for evaluation. The results of our model along with those of the state-of-the-art supervised models are reported in Table~\ref{tab::biwi_refinement}.
After refinement with the BIWI dataset's images, the MAE of SSV significantly reduces to \ang{5.8}. This demonstrates that SSV can improve its performance with the availability of images that match the target domain, even without GT annotations. We also show qualitative results of head pose estimation for this refined SSV-Full model in Figure~\ref{fig:pose_results}(a). It performs robustly to large variations in head pose, identity and expression.

\subsection{Generalization to Other Object Categories}
\vspace{-1mm}
SSV is not specific to faces and can be used to learn viewpoints of other object categories.
To demonstrate its generalization ability, we additionally train and evaluate SSV on the categories of cars, buses and trains.

\vspace{1mm}
\noindent \textbf{Datasets and evaluation metric}
Since SSV is completely self-supervised, the training image collection has to be reasonably large to cover all possible object viewpoints while covering diversity in other image aspects such as appearance, lighting etc. For this reason, we leverage large-scale image collections from both the existing datasets and the internet to train our network.
For the car category, we use the CompCars~\cite{yang2015large} dataset, which is a fine-grained car model classification dataset containing 137,000 car images in various viewpoints. 
For the `train' and `bus' categories, we use the OpenImages~\cite{papadopoulos2016we, papadopoulos2017extreme,benenson2019large} dataset which contains about 12,000 images of each of these categories. Additionally, we mine about 30,000 images from Google image search for each category. None of the aforementioned datasets have viewpoint annotations. 
This also demonstrates the ability of SSV to consume large-scale internet image collections that come without any viewpoint annotations.

We evaluate the performance of the trained SSV model on the test sets of the challenging Pascal3D+~\cite{xiang2014beyond} dataset. The images in this dataset 
have extreme shape, appearance and viewpoint variations. Following~\cite{mahendran20173d,prokudin2018deep,tulsiani2015viewpoints,LiaoCVPR19}, we estimate the azimuth, elevation and tilt values, given the GT object location. To compute the error between the predicted and GT viewpoints, we follow the standard geodesic distance $\Delta(R_{gt}, R_{p})=\left\lVert \log R^{T}_{gt}R_{p} \right\rVert_{\mathcal{F}}/\sqrt{2}$ between the predicted rotation matrix $R_{p}$ constructed using viewpoint predictions and $R_{gt}$ constructed using GT viewpoints~\cite{LiaoCVPR19}. 
Using this distance metric, we report the median geodesic error (Med. Error) for the test set. Additionally, we also compute the percentage of inlier predictions whose error is less than $\pi/6$ ($\mathrm{Acc}@\pi/6$).

\begin{figure*}[h]
	\centering
	\includegraphics[width=\linewidth,trim=0 .4cm 0 .6cm, clip]{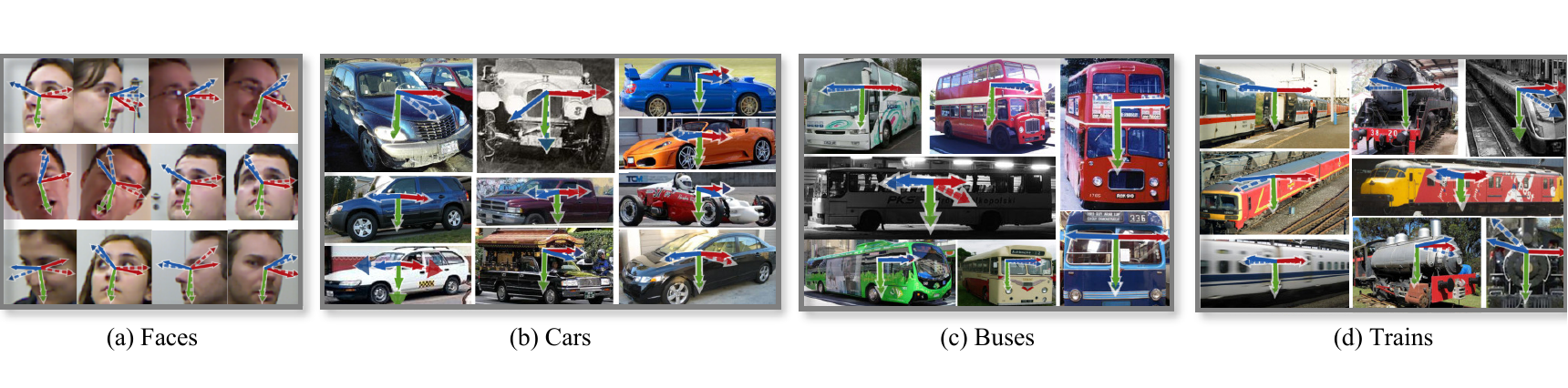}
	\mycaption{Viewpoint estimation results}{We visually show the results of (a) head pose estimation on the BIWI~\cite{fanelli2013random} dataset and of viewpoint estimation on the test sets of the (b) car, (c) bus and (d) train categories from the PASCAL3D+~\cite{xiang2014beyond} dataset. Solid arrows indicate predicted viewpoints, while the dashed arrows indicate their GT values. Our self-supervised method performs well for a wide range of head poses, identities and facial expressions. It also successfully handles different object appearances and lighting conditions from the car, bus and train categories. We show additional results in the supplementary material.\vspace{-2mm} 
	}
	\label{fig:pose_results}
\end{figure*}

\vspace{1mm}
\noindent \textbf{Baselines}
For head pose estimation, we compared with self-supervised landmark~\cite{zhang2018unsupervised,hung2019scops,tomas2018neurips} discovery techniques coupled with the PnP algorithm for head pose estimation by fitting them to an average 3D face. For objects like cars with full \ang{360} azimuth rotation, we notice that the landmarks produced by SCOPS~\cite{hung2019scops} and LMDIS~\cite{zhang2018unsupervised} cannot be used for reasonable viewpoint estimates. This is because SCOPS is primarily a self supervised part segmentation framework which does not distinguish between front and rear parts of the car. Since the keypoints we compute are the centers of part segments, the resulting keypoints cannot distinguish such parts. LMDIS on the other hand produces keypoints only for the side profiles of cars.
Hence, we use another baseline technique for comparisons on cars, trains and buses. Following the insights from \cite{hung2019scops,thewlisICCVT2017} that features learned by image classification networks are equivariant to object rotation, we 
learn a linear regressor that maps the Conv5 features of a pre-trained VGG network~\cite{Simonyan15_vgg} to the viewpoint of an object.
To train this baseline, we use the VGG image features and the GT viewpoint annotations in the Pascal3D+ training dataset~\cite{xiang2014beyond}. We use the same Pascal3D+ annotations used to calibrate SSV's predicted viewpoints to GT canonical viewpoint axes. We consider this as a self-supervised baseline since we are not using GT annotations for feature learning but only to map the features to viewpoint predictions. We refer to this baseline as VGG-View. As an additional baseline, we train  HoloGAN~\cite{nguyenhologan} with an additional viewpoint output and a corresponding loss for it. The viewpoint predictions are calibrated, similar to SSV.

\vspace{-1mm}
\noindent \textbf{Comparisons}
We compare SSV to our baselines and also to several state-of-the-art supervised viewpoint estimation methods on the Pascal3D+ test dataset. Table~\ref{tab::pvoc_mederrs} indicates that SSV significantly outperforms the baselines. With respect to supervised methods, SSV performs comparably to Tulsiani \etal~\cite{tulsiani2015viewpoints} and Mahendran \etal~\cite{mahendran20173d} in terms of Median error. Interestingly for the `train' category, SSV performs even better than supervised methods.
These results demonstrate the general applicability of SSV for viewpoint learning on different object categories. We show some qualitative results for these categories in Figure \ref{fig:pose_results}(b)-(d).

\setlength{\tabcolsep}{8pt} 
\begin{table}\centering
\begin{tabular}{rlccc}
\toprule
              & Method & Car & Bus & Train\\
\midrule
\small
\parbox[t]{0mm}{\multirow{3}{*}{\rotatebox[origin=c]{90}{Self-Sup}}}
\scriptsize
\rule{0pt}{2ex}
              
              & VGG-View                                    &  34.2 & 19.0   &   9.4 \\
              & HoloGAN~\cite{nguyenhologan} with $v$       &  16.3 & 14.2   &   9.7 \\
              \cdashline{2-5}
              \rule{0pt}{3ex} 
              & SSV-Full                             & \textbf{10.1} & \textbf{9.0}   & \textbf{5.3} \\
\midrule
\parbox[t]{1mm}{\multirow{4}{*}{\rotatebox[origin=c]{90}{Supervised}}}
              &Tulsiani \etal~\cite{tulsiani2015viewpoints}         & 9.1  & 5.8 & 8.7  \\
              &Mahendran \etal~\cite{mahendran20173d} & 8.1  & 4.3 & 7.3  \\
              &Liao \etal~\cite{LiaoCVPR19}                  & 5.2  & 3.4 & \textbf{6.1}  \\
              &Grabner \etal~\cite{grabner20183d}                   & \textbf{5.1}  & \textbf{3.3} & 6.7  \\
\midrule
\end{tabular}
\mycaption{Generalization to other object categories, median error} {We show the median geodesic errors (in degrees) for the car, bus and train categories.}
\label{tab::pvoc_mederrs}
\vspace{-1mm}
\end{table}

\setlength{\tabcolsep}{8pt} 
\begin{table}\centering
\begin{tabular}{rlccc}
\toprule
\rule{0pt}{1ex} 
              & Method & Car & Bus & Train\\
\midrule
\small
\parbox[t]{0mm}{\multirow{3}{*}{\rotatebox[origin=c]{90}{Self-sup}}}
\scriptsize
\rule{0pt}{2ex} 
              & VGG-View                                                      &  0.43    & 0.69  & 0.82  \\
              & HoloGAN~\cite{nguyenhologan} with $v$                         &  0.52    & 0.73  & 0.81  \\
             \cdashline{2-5}
             \rule{0pt}{3ex} 
              & SSV-Full                                                      & \textbf{0.67}    & \textbf{0.82} & \textbf{0.96}\\
\midrule
\parbox[t]{1mm}{\multirow{4}{*}{\rotatebox[origin=c]{90}{Supervised}}}
              &Tulsiani \etal~\cite{tulsiani2015viewpoints}                      & 0.89  &\textbf{ 0.98} & 0.80  \\
              &Mahendran \etal~\cite{mahendran20173d} & -    & -    & -      \\
              &Liao \etal~\cite{LiaoCVPR19}                               & \textbf{0.93}  & 0.97 & \textbf{0.84}   \\
              &Grabner \etal~\cite{grabner20183d}                                & \textbf{0.93}  & 0.97 & 0.80     \\
\midrule
\end{tabular}
\mycaption{Generalization to other object categories, inlier count} {We show the percentage of images with geodesic error less than $\pi/6$ for the car, bus and train categories.}
\label{tab::pvoc_inliers}
\vspace{-6mm}
\end{table}

%% file: 05_conclusions.tex
\vspace{-2mm}
\section{Conclusions}
\vspace{-2mm}
In this work we investigate the largely unexplored problem of learning viewpoint estimation in a self-supervised manner from collections of un-annotated object images. We design a viewpoint learning framework that receives supervision from a viewpoint-aware synthesis network; and from additional symmetry and adversarial constraints. We further supervise our synthesis network with additional losses to better control its image synthesis process. We show that our technique outperforms existing self-supervised techniques and performs competitively to fully-supervised ones on several object categories like faces, cars, buses and trains.

%% file: 06_supp.tex
\setcounter{section}{0}
\renewcommand{\thesection}{\Alph{section}}
\section*{\Large Appendix}

In this supplement, we provide the architectural and training details of our SSV framework. In Section \ref{sec::arch} we describe the architectures of the both viewpoint ($\mathcal{V}$) and synthesis ($\mathcal{S}$) networks. In Section \ref{sec::trn_details} we present the various training hyperparameters and the training schedule. In Section \ref{sec::mem_runtime} we examine the memory requirements and runtime of SSV. In Section \ref{sec::vis_results} we provide additional visual viewpoint estimation results for all object categories (\textit{i.e.}, face, car, bus and train). 

\section{Network Architecture}\label{sec::arch}
The network architectures of the viewpoint and synthesis networks are detailed in tables \ref{tab::vpnet_arch} and \ref{tab::synthnet_arch}, respectively. Both $\mathcal{V}$ and $\mathcal{S}$ operate at an image resolution of 128x128 pixels. $\mathcal{V}$ has an input size of 128x128. $\mathcal{S}$ synthesizes images at the same resolution. We use Instance Normalization~\cite{ulyanovcvpr17instancenorm} in the viewpoint network. For the synthesis network, the size of the style code $\boldsymbol{z}_s$ is 128 for faces and 200 for the other objects (car, bus and train). $\boldsymbol{z}_s$ is mapped to affine transformation parameters ($\gamma(\boldsymbol{z}_s), \sigma(\boldsymbol{z}_s)$), which are in turn used by adaptive instance normalization(AdaIN)~\cite{huang2017arbitrary} to control the style of the synthesized images.

\section{Training Details}\label{sec::trn_details}
SSV is implemented in Pytorch~\cite{paszke2017automatic}. We open-source our code required to reproduce the results at \url{https://github.com/NVlabs/SSV}. We train both our viewpoint and synthesis networks from scratch by initializing all weights with a normal distribution $\mathcal{N}(0,0.2)$ and zero bias. The learning rate is 0.0001 for both ($\mathcal{V}$) and ($\mathcal{S}$). We use the ADAM~\cite{kingma2015adam} optimizer with betas (0.9, 0.99) and no weight decay. We train the networks for 20 epochs. 

\vspace{2mm}
\noindent \textbf{Training Cycle}~
In each training iteration, we optimize $\mathcal{V}$ and $\mathcal{S}$ alternatively. In the $\mathcal{V}$ optimization step, we compute the generative consistency, discriminator loss and the symmetry constraint (Sections 3.1, 3.2, 3.3 in the main paper). We freeze the parameters of $\mathcal{S}$, compute the gradients of the losses with respect to parameters of $\mathcal{V}$ and do an update step for it. In an alternative step, while optimizing $\mathcal{S}$, we compute the paired style and viewpoint consistency, flip image consistency and the adversarial loss (Section 4 in the paper). We freeze the parameters of $\mathcal{V}$, compute the gradients of the losses with respect to parameters of $\mathcal{S}$ and do an update step for it. We train separate networks for each object category. 

\section{Runtime and Memory}\label{sec::mem_runtime}
Our viewpoint network $\mathcal{V}$ runs real-time with 76 FPS. That is, the inference takes 13 milliseconds 
on an NVIDIA Titan X Pascal GPU for a single image.
The memory consumed is 900MB. 
We use a small network for viewpoint estimation for real-time performance
and low-memory consumption. 

\section{Visual Results}\label{sec::vis_results}
\begin{figure*}[h]
	\centering
	\includegraphics[width=\linewidth]{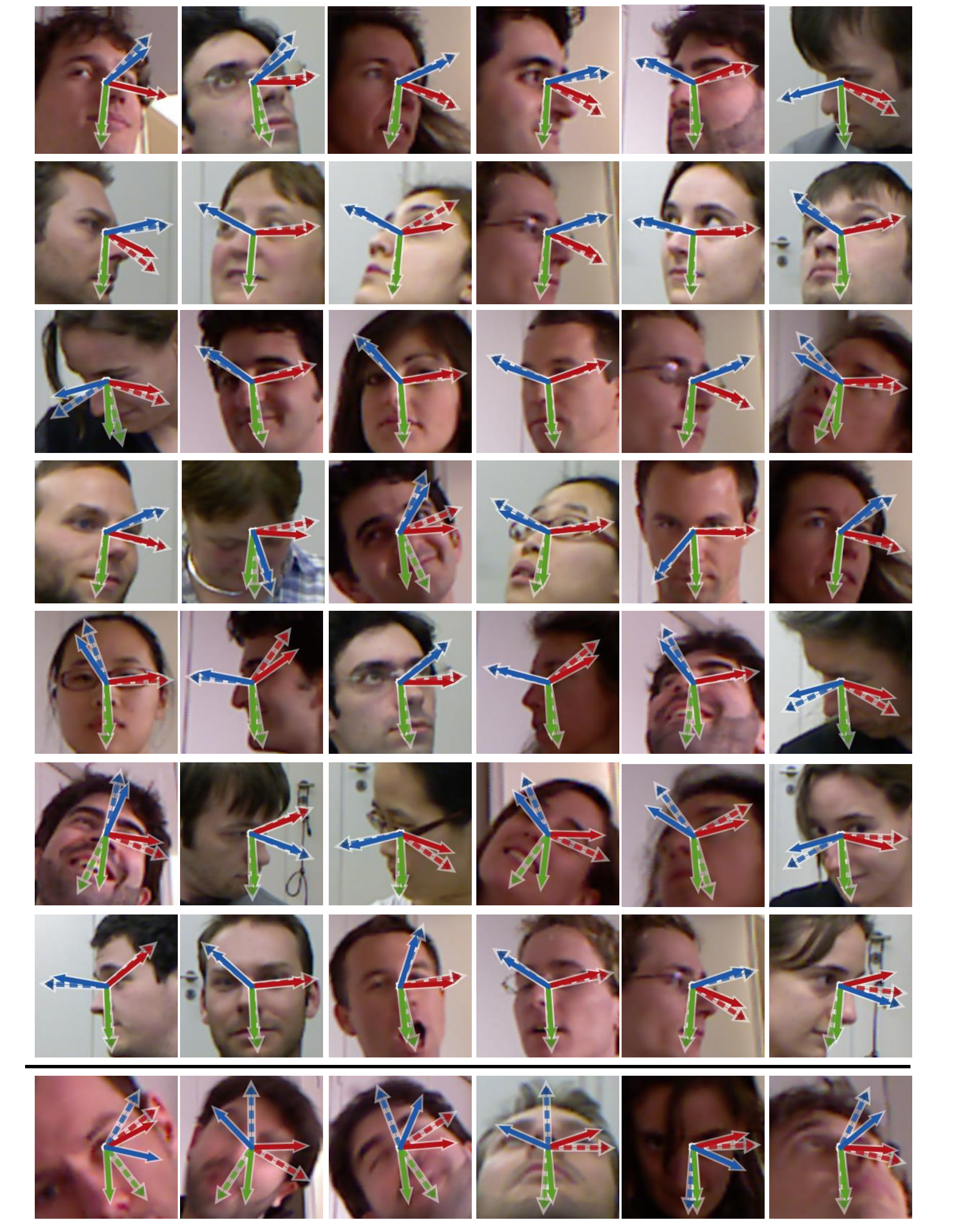}
	\mycaption{Viewpoint estimation results for the face category}{SSV predicts reliable viewpoints for a variety of face poses with large variations in azimuth, elevation and tilt. The last row (below the black line) shows some erroneous cases where the faces are partially detected by the face detector or there are extreme elevation angles.}
	\label{fig:sup_face_pose_results}
\end{figure*}

\begin{figure*}[h]
	\centering
	\includegraphics[width=\linewidth]{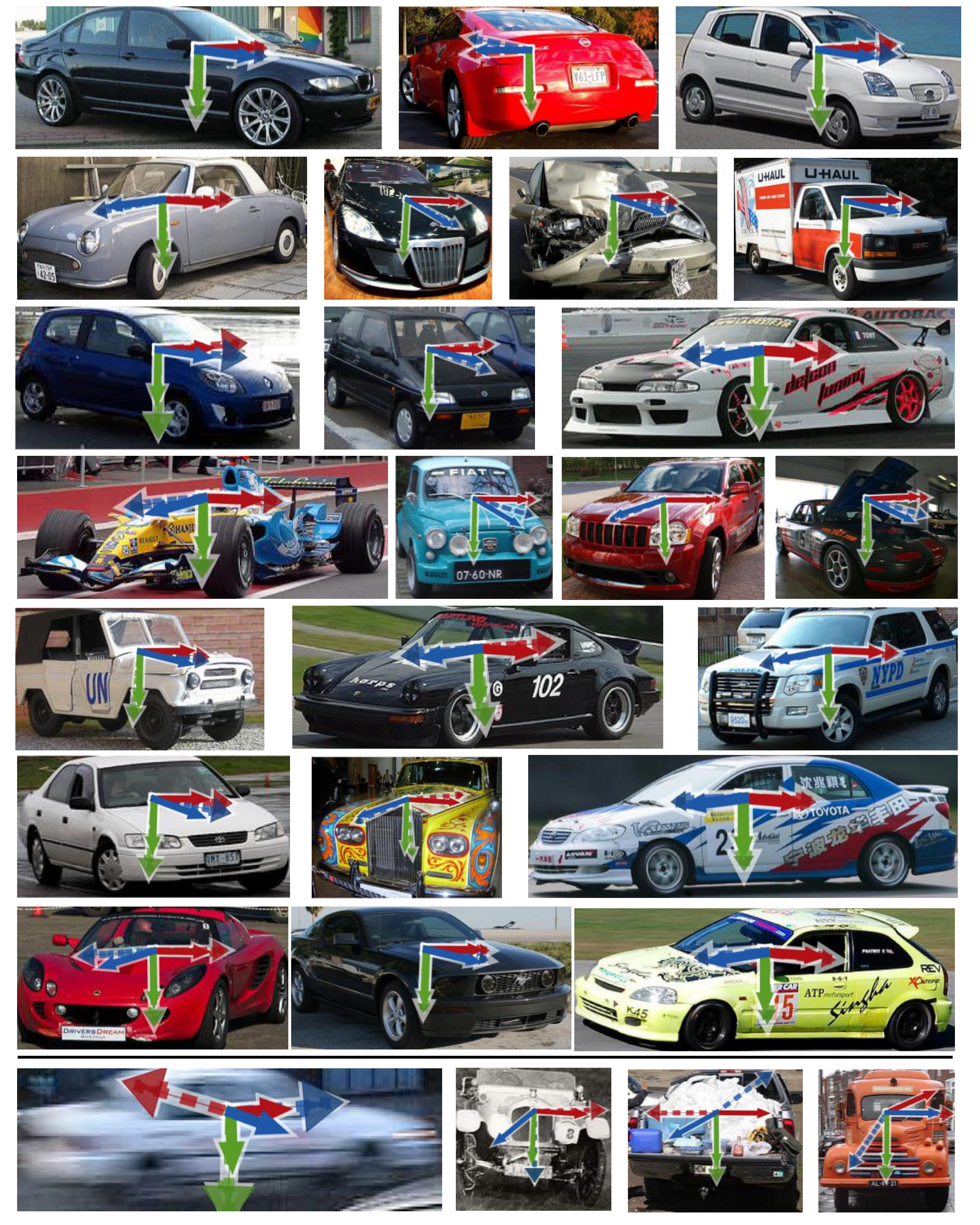}
	\mycaption{Viewpoint estimation results for the car category}{SSV predicts reliable viewpoints for a variety of objects with large variations in azimuth, elevation and tilt. It generalizes to car models like race cars and formula-1 cars, which are not seen by SSV during training. The last row (below the black line) shows some erroneous cases where the objects have extreme motion blur or are heavily occluded to the extent where it is difficult to identify it as a car. }
	\label{fig:sup_car_pose_results}
\end{figure*}

In figures \ref{fig:sup_face_pose_results}, \ref{fig:sup_car_pose_results}, \ref{fig:sup_bus_pose_results}, we present some additional visual results for the various object categories (faces, cars, buses and trains). It can be seen that the viewpoint estimation network reliably predicts viewpoint. For cars, it generalizes to car models like race cars and formula-1 cars, which are not seen by SSV during training. In each figure, we also show some failure cases in the last row. For faces, We observe that failures are caused in cases where the viewpoints contain extreme elevation or noisy face detection. For cars, viewpoint estimation is noisy when there is extreme blur in the image or the if the car is heavily occluded to the extent where it is difficult to identify it as a car. For buses, viewpoint estimation is erroneous when there is ambiguity between the rear and front parts of the object. 

\setlength{\tabcolsep}{8pt} 
\begin{table*}[h]
	\begin{tabular}{ccccccc}
		\toprule
		& Layer & Kernel Size & stride & Activation & Normalization & Output Dimension \\ 
		\midrule
		& Conv  & 	1x1 	&	1	 & 	 LReLU	  &		  -    	  & 128x128x128   \\ 
		\midrule
		
		\parbox[t]{2mm}{\multirow{17}{*}{\rotatebox[origin=c]{90}{Backbone Layers}}}		
		
		& Conv2D  & 	3x3 	&	1	 & 	 LReLU	  &	Instance Norm & 128X128x256   \\
		& Conv2D  & 	3x3 	&	1	 & 	 LReLU	  &	Instance Norm & 128X128x256   \\
		\cdashline{2-7}
		\rule{0pt}{3ex}
		& &	  &				&	 Interpolate (scale = 0.5) 	& 		  &				\\
		\cdashline{2-7}
		\rule{0pt}{3ex}
		
		& Conv2D & 	3x3 	&	1	 & 	 LReLU	  &	Instance Norm & 64X64x512   \\				
		& Conv2D & 	3x3 	&	1	 & 	 LReLU	  &	Instance Norm & 64X64x512   \\				
		\cdashline{2-7}
		\rule{0pt}{3ex}
		& &	  &				&	 Interpolate (scale = 0.5) 	& 		  &				\\
		\cdashline{2-7}	
		\rule{0pt}{3ex}	
		
		& Conv2D & 	3x3 	&	1	 & 	 LReLU	  &	Instance Norm & 32X32x512   \\				
		& Conv2D & 	3x3 	&	1	 & 	 LReLU	  &	Instance Norm & 32X32x512   \\				
		\cdashline{2-7}
		\rule{0pt}{3ex}
		& &	  &				&	 Interpolate (scale = 0.5) 	& 		  &				\\
		\cdashline{2-7}
		\rule{0pt}{3ex}
		
		& Conv2D & 	3x3 	&	1	 & 	 LReLU	  &	Instance Norm & 16X16x512   \\				
		& Conv2D & 	3x3 	&	1	 & 	 LReLU	  &	Instance Norm & 16X16x512   \\				
		\cdashline{2-7}
		\rule{0pt}{3ex}
		& &	  &				&	 Interpolate (scale = 0.5) 	& 		  &				\\
		\cdashline{2-7}
		\rule{0pt}{3ex}
		& Conv2D & 	3x3 	&	1	 & 	 LReLU	  &	Instance Norm & 8X8x512   \\				
		& Conv2D & 	3x3 	&	1	 & 	 LReLU	  &	Instance Norm & 8X8x512   \\				
		\cdashline{2-7}
		\rule{0pt}{3ex}
		& &	  &				&	 Interpolate (scale = 0.5) 	& 		  &			\\
		\cdashline{2-7}
		\rule{0pt}{3ex}
		
		& Conv2D & 	3x3 	&	1	 & 	 LReLU	  &	Instance Norm & 4X4x512   \\				
		& Conv2D & 	4x4 	&	1	 & 	 LReLU	  &		-  & 1X1x512   \\				
		\cdashline{2-7}
		\rule{0pt}{3ex}
		& &	  &				&	 Backbone ouput 	& 		  &			\\
		\midrule
		\midrule
		& FC-real/fake  & 	- 	&	-	 & 	 -	  &		-  & 1 \\				
		\midrule
		\midrule		
		& FC-style  & 	- 	&	-	 & 	 -	  &		-  &  code\_dim\\				
		\midrule		
		\midrule
		\parbox[t]{2mm}{\multirow{3}{*}{\rotatebox[origin=c]{90}{Azimuth}}}
		& FC  & 	- 	&	-	 & 	 LReLU	  &		-  & 256 					\\				
		& FC - $\lvert \hat{a} \rvert$  & 	- 	&	-	 & 	   -	  &		-  & 2  					\\				
		& FC - sign($\hat{a}$) & 	- 	&	-	 & 	   -	  &		-  & 4  					\\				
		\midrule		
		\midrule
		\parbox[t]{2mm}{\multirow{3}{*}{\rotatebox[origin=c]{90}{Elevation}}}		
		& FC  & 	- 	&	-	 & 	 LReLU	  &		-  & 256 					\\				
		& FC - $\lvert \hat{e} \rvert$  & 	- 	&	-	 & 	   -	  &		-  & 2  					\\				
		& FC - sign($\hat{e}$) & 	- 	&	-	 & 	   -	  &		-  & 4  					\\				
		\midrule		
		\midrule
		\parbox[t]{2mm}{\multirow{3}{*}{\rotatebox[origin=c]{90}{Tilt}}}				
		& FC  & 	- 	&	-	 & 	 LReLU	  &		-  & 256 					\\				
		& FC - $\lvert \hat{e} \rvert$  & 	- 	&	-	 & 	   -	  &		-  & 2  					\\				
		& FC - sign($\hat{e}$) & 	- 	&	-	 & 	   -	  &		-  & 4  					\\				
		\midrule		
		
	\end{tabular}
	\mycaption{Viewpoint Network Architecture}{The network contains a backbone whose resultant fully-connected features are shared by the heads that predict (a) real/fake scores, (b) style codes, and (c) heads that predict azimuth, elevation and tilt values. All LReLU units have a slope of 0.2. FC indicates a fully connected layer.  }
	\label{tab::vpnet_arch}
\end{table*}

\setlength{\tabcolsep}{8pt} 
\begin{table*}[h]
	\begin{tabular}{rcccccc}
		\toprule
		& Layer & Kernel Size & stride & Activation & Normalization & Output Dimension \\ 
		\midrule
		& Input - 3D Code  	  &  - 	   &	-	 	& 	 -	  		&		  -    	  & 4x4x4x512   \\ 
		\midrule
		\parbox[t]{4mm}{\multirow{8}{*}{\rotatebox[origin=c]{90}{Styled 3D Convs}}}
		& Conv 3D  			  &  3x3   &  	1 		&	LReLU 		& 	 AdaIN 		&	4x4x4x512   \\ 
		& Conv 3D  			  &  3x3   &  	1 		&	LReLU 		& 	 AdaIN 		&	4x4x4x512   \\ 
		\cdashline{2-7}
		\rule{0pt}{3ex}		
		& &	  &				  &	 Interpolate (scale = 2) 	& 		  &				\\
		\cdashline{2-7}	
		\rule{0pt}{3ex}		
		& Conv 3D  			  &  3x3   &  	1 		&	LReLU 		& 	 AdaIN 		&	8x8x8x512   \\ 
		& Conv 3D  			  &  3x3   &  	1 		&	LReLU 		& 	 AdaIN 		&	8x8x8x512   \\ 
		\cdashline{2-7}
		\rule{0pt}{3ex}		
		& &	  &				  &	 Interpolate (scale = 2) 	& 		  &				\\
		\cdashline{2-7}	
		\rule{0pt}{3ex}		
		& Conv 3D  			  &  3x3   &  	1 		&	LReLU 		& 	 AdaIN 		&	16x16x16x256   \\ 
		& Conv 3D  			  &  3x3   &  	1 		&	LReLU 		& 	 AdaIN 		&	16x16x16x256   \\ 
		\midrule
		& &	  &	              &	 3D Rotation  		& 		  		&				\\
		\midrule
		& Conv 3D  			  &  3x3   &  	1 		&	LReLU 		& 	 - 		&	16x16x16x128   	\\ 
		& Conv 3D  			  &  3x3   &  	1 		&	LReLU 		& 	 - 		&	16x16x16x128   	\\ 
		\cdashline{2-7}	
		\rule{0pt}{3ex}		
		& Conv 3D  			  &  3x3   &  	1 		&	LReLU 		& 	 - 		&	16x16x16x64   	\\ 
		& Conv 3D  			  &  3x3   &  	1 		&	LReLU 		& 	 - 		&	16x16x16x64   	\\ 
		\midrule
		\rule{0pt}{3ex}				
		\parbox[t]{4mm}{\multirow{2}{*}{\rotatebox[origin=c]{90}{Project}}}		
		& Collapse 			  &  -	   &  	- 		&	-	 		& 	 - 		&	16x16x(16.64)   \\ 
		& Conv 			  	  &  3x3   &  	1 		&	LReLU	    & 	 - 		&	16x16x1024   	\\ 
		\small
		&  			  	  &     &  	 		&		    & 	  		&	   	\\ 
		\midrule
		\parbox[t]{4mm}{\multirow{11}{*}{\rotatebox[origin=c]{90}{Styled 2D Convs}}}		
		& Conv 2D  			  &  3x3   &  	1 		&	LReLU 		& 	 AdaIN  &	16x16x512   \\ 
		& Conv 2D  			  &  3x3   &  	1 		&	LReLU 		& 	 AdaIN 	&	16x16x512   \\ 
		\cdashline{2-7}	
		\rule{0pt}{3ex}		
		& &	  &				  &	 Interpolate (scale = 2) 	& 		  &				\\
		\cdashline{2-7}	
		\rule{0pt}{3ex}		
		& Conv 2D  			  &  3x3   &  	1 		&	LReLU 		& 	 AdaIN  &	32x32x256   \\ 
		& Conv 2D  			  &  3x3   &  	1 		&	LReLU 		& 	 AdaIN 	&	32x32x256   \\ 
		\cdashline{2-7}	
		\rule{0pt}{3ex}		
		& &	  &				  &	 Interpolate (scale = 2) 	& 		  &				\\
		\cdashline{2-7}	
		\rule{0pt}{3ex}				
		& Conv 2D  			  &  3x3   &  	1 		&	LReLU 		& 	 AdaIN  &	64x64x128   \\ 
		& Conv 2D  			  &  3x3   &  	1 		&	LReLU 		& 	 AdaIN 	&	64x64x128   \\ 
		\cdashline{2-7}	
		\rule{0pt}{3ex}		
		& &	  &				  &	 Interpolate (scale = 2) 	& 		  &				\\
		\cdashline{2-7}	
		\rule{0pt}{3ex}		
		& Conv 2D  			  &  3x3   &  	1 		&	LReLU 		& 	 AdaIN  &	128x128x64   \\ 
		& Conv 2D  			  &  3x3   &  	1 		&	LReLU 		& 	 AdaIN 	&	128x128x64   \\ 
		\midrule		
		Out & Conv 2D  		  &  3x3   &  	1 		&	 - 			& 	  - 	&	128x128x3   \\ 
		\midrule		
		
	\end{tabular}
	\mycaption{Synthesis Network Architecture}{This network contains a set of 3D and 2D convolutional blocks. A learnable 3D latent code is passed through stylized 3D convolution blocks, which also use style codes as inputs to their adaptive instance normalization(AdaIN~\cite{huang2017arbitrary}) layers. The resulting 3D features are then rotated using a rigid rotation via the input viewpoint. Following this, the 3D features are orthographically projected to become 2D features. These are then passed through a stylized 2D convolution network which has adaptive instance normalization layers to control the style of the synthesized image. }
	\label{tab::synthnet_arch}
\end{table*}

\begin{figure*}[h]
	\centering
	\includegraphics[width=\linewidth]{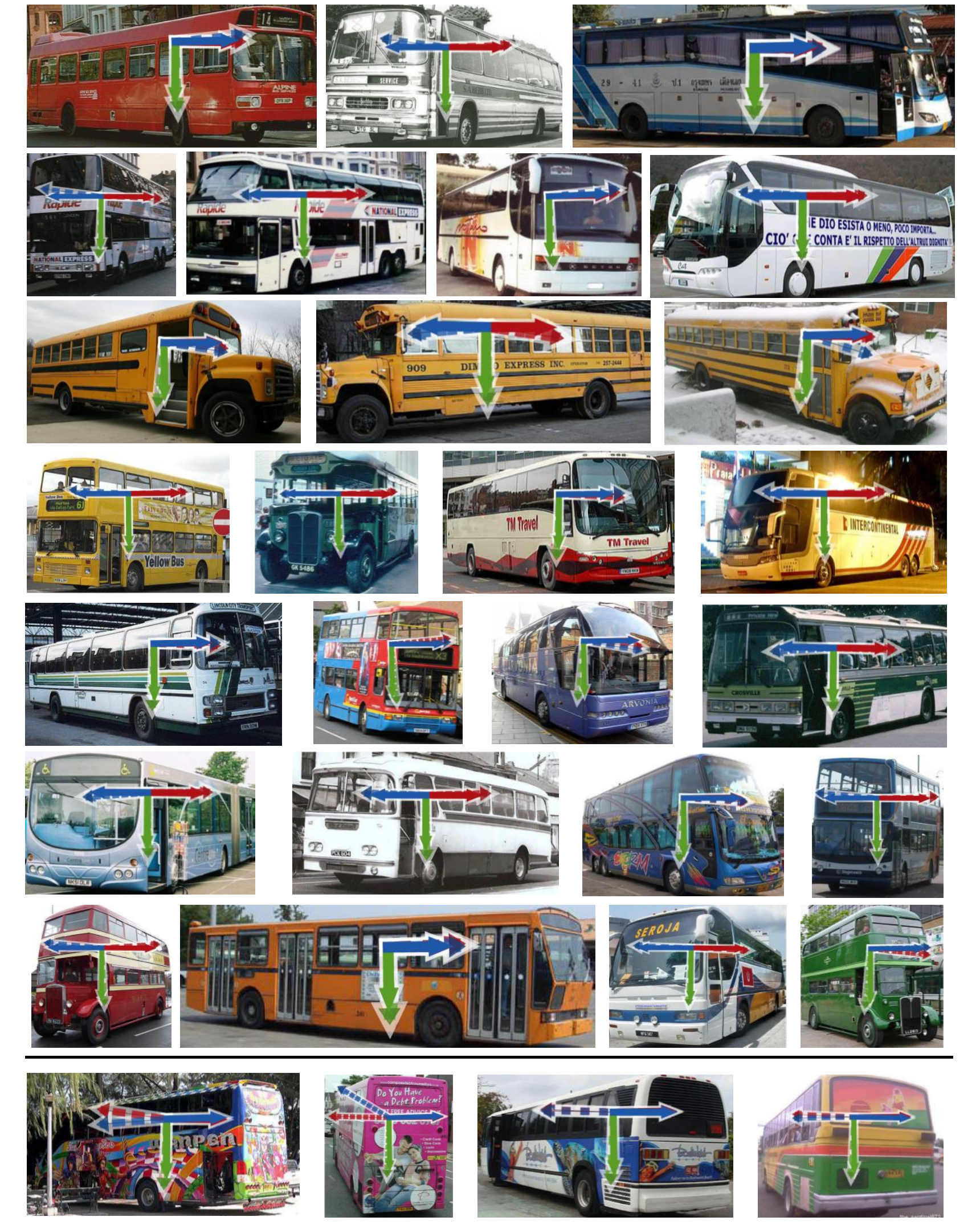}
	\mycaption{Viewpoint estimation results for the bus category }{SSV predicts reliable viewpoints for a variety of buses with large variations in azimuth, elevation and tilt. The last row (below the black line) shows erroneous viewpoints when there is ambiguity between the rear and front parts of the object.}
	\label{fig:sup_bus_pose_results}
\end{figure*}

\begin{figure*}[h]
	\centering
	\includegraphics[width=\linewidth]{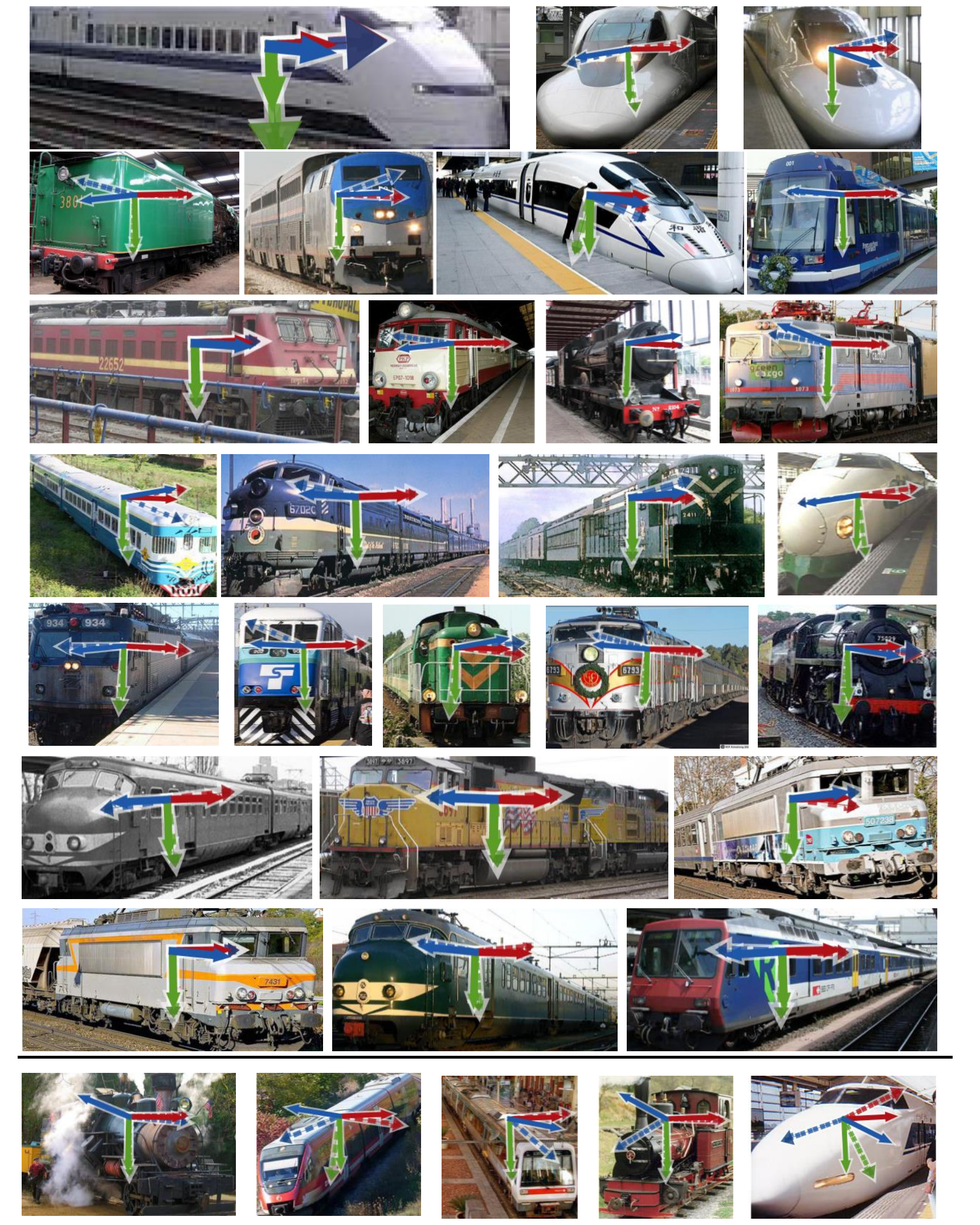}
	\mycaption{Viewpoint estimation results for the train category}{SSV predicts reliable viewpoints for a variety of objects with large variations in azimuth, elevation and tilt. The last row (below the black line) shows the erroneous viewpoints predicted by SSV.}
	\label{fig:sup_bus_pose_results}
\end{figure*}